\documentclass[letterpaper, 10 pt, conference]{ieeeconf}  

\IEEEoverridecommandlockouts                              
\usepackage{graphicx}
\usepackage{float}
\usepackage{caption}
\usepackage{subcaption}
\usepackage{multicol}
\usepackage{multirow}
\usepackage{hyperref}

\usepackage{amsmath,bm}
\usepackage{amssymb}
\usepackage{indentfirst}
\usepackage{graphicx}
\usepackage{wrapfig}
\usepackage{color,soul} 
\newcommand{\scalemath}[2]{\scalebox{#1}{\mbox{\ensuremath{\displaystyle #2}}}}
\definecolor{red}{rgb}{1.00,0.00,0.00}
\definecolor{blue}{rgb}{0.00,0.00,1.00}
\definecolor{green}{rgb}{0.30, 0.50,0.00}

\newcommand{\cblue}[1] {\textcolor{blue}{#1}}

\usepackage{flushend} 

\title{\LARGE \bf Throwing Objects into A Moving Basket While Avoiding Obstacles
}

\author{Hamidreza Kasaei$^{1}$ and Mohammadreza Kasaei$^{2}$
\thanks{$^{1}$Hamidreza~Kasaei is with the Department of Artificial Intelligence, Bernoulli Institute, Faculty of Science and Engineering, University of Groningen, The Netherlands. Email: hamidreza.kasaei@rug.nl}%
\thanks{$^{2}$ Mohammadreza Kasaei is with the School of Informatics, University of Edinburgh, UK. Email: m.kasaei@ed.ac.uk}%
}

\begin{document}

\maketitle
\thispagestyle{empty}
\pagestyle{empty}

\begin{abstract}

The capabilities of a robot will be increased significantly by exploiting \textit{throwing} behavior. In particular, throwing will enable robots to rapidly place the object into the target basket, located outside its feasible kinematic space, without traveling to the desired location. In previous approaches, the robot often learned a parameterized throwing kernel through analytical approaches, imitation learning, or hand-coding. There are many situations in which such approaches do not work/generalize well due to various object shapes, heterogeneous mass distribution, and also obstacles that might be presented in the environment. It is obvious that a method is needed to modulate the throwing kernel through its meta-parameters. In this paper, we tackle object throwing problem through a deep reinforcement learning approach that enables robots to precisely throw objects into moving baskets while there are obstacles obstructing the path. To the best of our knowledge, we are the first group that addresses throwing objects with obstacle avoidance. Such a throwing skill not only increases the physical reachability of a robot arm but also improves the execution time. In particular, the robot detects the pose of the target object, basket, and obstacle at each time step, predicts the proper grasp configuration for the target object, and then infers appropriate parameters to throw the object into the basket. Due to safety constraints, we develop a simulation environment in Gazebo to train the robot and then use the learned policy in real-robot directly. To assess the performers of the proposed approach, we perform extensive sets of experiments in both simulation and real robots in three scenarios. Experimental results showed that the robot could precisely throw a target object into the basket outside its kinematic range and generalize well to new locations and objects without colliding with obstacles. The video of our experiments can be found at \href{https://youtu.be/VmIFF__c_84}{\cblue{https://youtu.be/VmIFF\_\_c\_84}}

\end{abstract}
\section{Introduction}

Almost all humans are familiar with the ability to throw objects, as we learn how to throw a ball during a game (e.g., basketball) or an object into a bin (e.g., tossing dirty clothes into the laundry basket). We throw objects either to speed up tasks by reducing the time of pick-and-place or to place them in an unreachable place~\cite{kuhn2016throwing}. Therefore, adding such a throwing motion to a robotic manipulator would enhance its functionality too. In particular, throwing object is a great way to use dynamics and increase the power of a robot by enabling it to quickly place objects into the target locations outside of the robot's kinematic range. However, the act of precisely throwing is actually far more complex than it appears and requires a lot of practice since it depends on many factors, ranging from pre-throw conditions (e.g. initial pose of the object inside the gripper) to the physical properties of the object (e.g. shape, size, softness, mass, material, etc.). Many of these elements are challenging to describe or measure analytically, hence, earlier research has frequently been limited to assuming predefined objects (e.g., ball) and initial conditions (e.g., manually placing objects in a per-defined location). When obstacles are present in the environment and the target basket is moving, throwing becomes even more difficult. To the best of our knowledge, we are the first group to address such a challenging object throwing problem.

To accomplish the throwing task successfully, a robot must process the visual information to realize which objects exist in the scene (i.e., target object, basket, and obstacles), what are the state of the objects (i.e., pose, speed, etc.), and how to grasp the target object (grasp synthesis). The robot then finds an obstacle-free trajectory to grasp the object. Afterward, given the obstacles that exist in the scene and the state of the target basket, it needs to predict throwing parameters to throw the object to the desired location precisely (e.g., the velocity of executing the throw trajectory, time of release, etc.). Lastly, the robot executes the throwing motion using those parameters. 

\begin{figure}[!t]
    \centering
\includegraphics[width=0.95\linewidth]{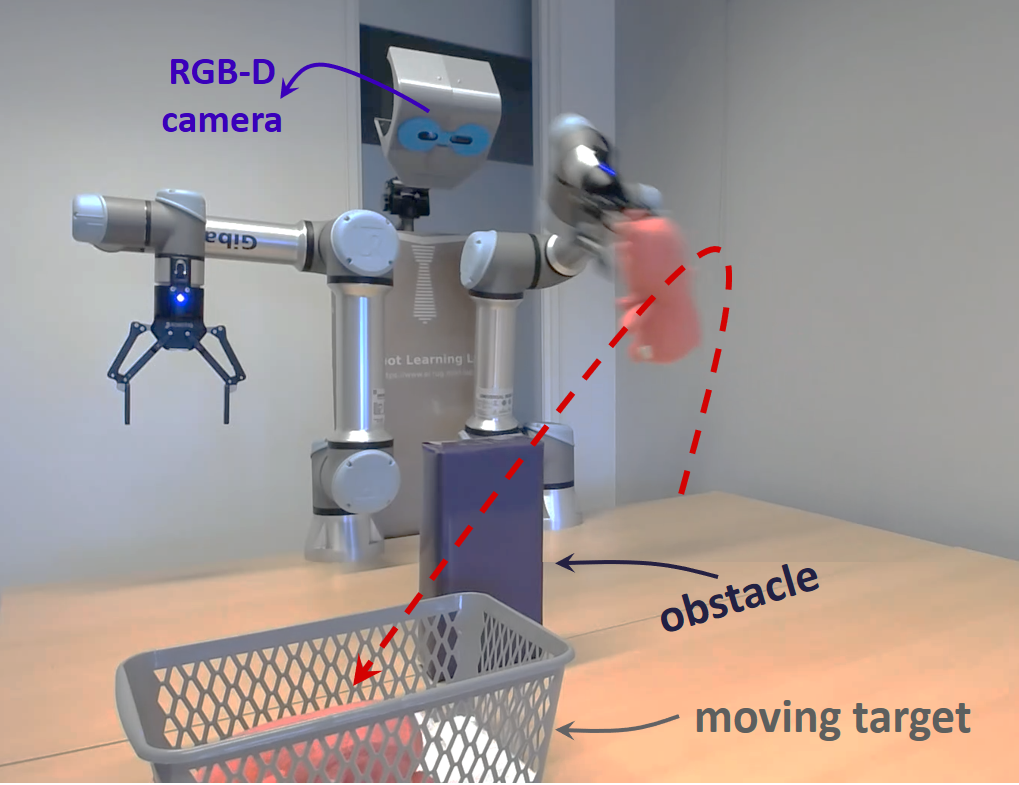}
\caption{An example scenario of throwing an object into a moving basket located outside of the robot's workspace while an obstacle obstructing the path. To accomplish this task successfully, the robot should perceive the environment through its RGB-D camera, and then infer the proper parameters to throw the object into the basket.}
\label{fig:obstacle_avoidance}
\vspace{-4mm}
\end{figure}

In this paper, we formulate object throwing as a RL problem to enable the robot to generalize well across a variety of objects and react quickly to dynamic environments (i.e., moving basket). For RL, the exploration phase is often unsafe in the real-world. It takes a while to build up enough experience to train the policy to function successfully in a dynamic environment with moving targets and obstacles. Therefore, we develop a simulation in Gazebo, very similar to our real-robot setup, and train the robot in Gazebo initially. Afterwards, the learned policy is used in real-robot settings directly. We extensively evaluate the performance of our approach in both simulation and real-robot using three different tasks with ascending levels of difficulties. Experimental results show that the proposed method produces throws that are more accurate than baseline alternatives. In summary, our key contributions are threefold:

\begin{itemize}
    \item To the best of our knowledge, we are the first group that addresses object tossing while obstacles are present in the environment and the target basket is moving.

    \item Despite only trained using simulation data, the proposed approach can be directly applied to real-robot. Furthermore, it shows impressive generalization capability to new target locations and unseen objects. 

    \item Our experiments show that the trained policy could achieve above $80\%$ object throwing accuracy for the most difficult task (i.e., throwing object into the basket while there is an obstacle obstructing the path) in both simulation and real robot environments.
    
\end{itemize}

\section{Related Work}
The robotics community has long been interested in giving service robots the ability to throw objects~\cite{hu2010ball,gai2013motion,zeng2020tossingbot,takahashi2021dynamic,ploeger2022controlling}. Throwing formulae were mostly influenced by analytical models in the late 1990s and early 2000s \cite{mason1993dynamic}, while such formulations are increasingly moving toward learning approaches today~\cite{kober2012learning,zeng2020tossingbot}. In the following subsections, we briefly review these approaches. 

\subsection{Analytical Approaches}

Earlier throwing systems relied on handcrafting or mechanical analysis and then optimizing control parameters to execute a throw such that the projectile (typically a ball) lands at a target location. As we previously highlighted, precisely modeling of dynamics is difficult because it calls for knowledge about the physical characteristics of the object, gripper and environment, which are hard to quantify~\cite{mason1993dynamic}. For instance, Y. Gai et al., derived an analytical approach for throwing a ball using a manipulator with a single flexible link through Hamilton’s principle~\cite{gai2013motion}. This is an example of tuning for a single object, a ball in this case. In another work, Jwu-Sheng Hu et al.,~\cite{hu2010ball} discussed a stereo vision system for throwing a ball into a basket. They calculated the ball-throwing transformation for a specific ball object based on cubic polynomial. In \cite{lofaro2012humanoid}, an analytical approach is used to predict the end-effector velocity (magnitude and direction) as well as a duration movement for underhand throwing task by a humanoid robot. Such approaches to some extend work for specific scenario but have difficulties generalizing over changing dynamics and various objects. 

\subsection{Learning Approaches}

Unlike analytical approaches for throwing, learning-based methods enable robots to learn/optimize the main task directly through success or failure signals. In general, learning-based throwing approaches demonstrate better performance than analytical methods\cite{ghadirzadehdeep,kober2011reinforcement}. In \cite{ghadirzadehdeep}, a deep predictive policy training architecture (DPPT) is presented to teach a PR2 robot object-grasping and ball-throwing tasks. They showed DPPT is successful in both simulated and real robots. In another work, Kober et al.~\cite{kober2011reinforcement} introduced an RL-based method for dart throwing task based on a kernelized version of the reward-weighted regression. In both of these works, the properties of the object (ball and dart) are known a-priori. In contrast to both of these approaches, we do not make assumptions about the physical properties of objects that are thrown.

In some other works, researchers tried to combine the potential of analytical and learning approaches for robotic throwing tasks. In particular, analytical models are used to approximate the initial control parameters, and a learning-based model is used to estimate residual parameters to adjust the initial parameters. Such approaches are called residual physics. For instance, \cite{zeng2020tossingbot} proposed TossingBot, an end-to-end self-supervised learning method for learning to throw arbitrary objects with residual physics. Similar to our work, their approach was able to throw an object into a basket. Unlike our approach, they used an analytical approach for estimating initial control parameters, and then used an end-to-end formulation for learning residual velocity for throwing motion primitives. We formulate the throwing task as an RL problem that modulates the parameters of a kernel motion generator. In contrast to all reviewed works, our formulation allows the robot to throw the object into a moving basket while avoiding present obstacles, whereas, in all reviewed works, the throwing task is considered in an obstacle-free environment where the target is static and known in advance.  
\section{Method} 
\label{problemformulation}

In this section, the preliminaries are briefly reviewed, followed by a discussion of how we formulate object throwing as an RL problem. The perception that represents the world model at each time step is the subject of the last subsection.

\subsection{Preliminaries}
\textbf{Markov Decision Process (MDP):} An MDP can be described as a tuple containing four basic elements: $(s_{t}, a_{t}, p(s_{t+1}|s_{t}, a_{t}), r(s_{t+1}|s_{t}, a_{t}))$, where the $s_t$ and $a_t$ are the continuous state and action at time step $t$, respectively. $p(s_{t+1}|s_{t}, a_{t})$ shows the transition probability function to reach to the next state $s_{t+1}$ given the current state $s_{t}$ and action $a_t$. The $r(s_{t+1}|s_{t}, a_{t})$ denotes the immediate reward received from the environment after the state transition. 

\textbf{Off-policy RL:} In online RL, an agent continuously interacts with the environment to accumulate experiences for learning the optimal policy$\pi^*$. The agent seeks to maximize the expected future return $R_{t}=\mathbb{E}[\sum_{i=t}^\infty\gamma^{i-t}r_{i+1}]$ with a discounted factor $\gamma \in [0, 1]$ weighting the future importance.  The expected return under a policy $\pi$ after taking action $a$ in the state  $s$ is computed by a corresponding action-value function $Q^{\pi}(s, a)=\mathbb{E}[R_t|s_t=s, a_t=a]$. By following policy $\pi$, we can compute $Q^{\pi}$ through the Bellman equation:
\begin{equation}\label{Qfunc}
\begin{split}
\scalemath{0.87}{
Q^{\pi}(s_t, a_t)  = \mathbb{E}_{s_{t+1}\sim p}[r(s_t, a_t)+ \gamma\mathbb{E}_{a_{t+1}\sim{A}}[Q_{\pi}(s_{t+1}, a_{t+1})]],
}
\end{split}
\end{equation}
where $A$ states the action space. Consider $Q^{*}(s, a)$ is the optimal action-value function. RL algorithms aim to find an optimal policy $\pi^*$ such that $Q^{\pi^*}(s, a)=Q^{*}(s, a)$ for all states and actions. 

\subsection{Problem Formulation}
Given the start pose~$p_s$ (i.e., grasp synthesis of an object) and the goal pose~$p_g$ (i.e., pose of the basket), the throwing task is defined as the problem of finding proper parameters to throw the object into the basket while avoiding obstacles~$o$ in the environment. A fully observable MDP can be used to represent this task, and the off-policy reinforcement learning framework can be used to solve it. We discuss the detailed RL formulation of throwing task in the following subsections.

\subsubsection{\textbf{States}}


A feature vector is used to describe the continuous state, including the robot's proprioception, the pose of the obstacle and the goal in the environment, releasing time, duration of trajectory execution, and the distance between the thrown object and the goal. In particular, we formed a kernel trajectory for throwing an object in a straight way using trial and error. We then let the robot learns the initial and final value for the shoulder joint, speed of executing the trajectory, and releasing time in the learning process. 

In particular, at each step $t$, we record the initial and final shoulder joint values ($j_i$ and $j_f$) in radians as the proprioception: ${\rm proprio}=(j_i, j_f) \in \mathbb{R}^2$. Then, we estimate the obstacle's position in task space and describe it as ${\rm obs} = (x^{o}, y^{o}, z^{o}) \in \mathbb{R}^3$. We describe the position of goal (i.e, center of the box) as a point in the task space: ${\rm goal}=(x^g, y^g, z^g, \dot{x}^g, \dot{y}^g, \dot{z}^g) \in \mathbb{R}^6$. We also consider the absolute distance of the thrown object relative to the obstacle and goal, and also the distances in the $X$ and $Y$ axes, ${\rm dist}=(d^g, d^g_x, d^g_y, d^o, d^o_x, d^o_y) \in \mathbb{R}^6$. We also record two timing profiles, the duration of executing the throwing trajectory $\tau$, and the time for releasing the object, $t_r$, where $t_r < \tau$ and ${\rm time} = (t_r , \tau) \in \mathbb{R}^2$. Finally, the state feature vector can be represented as: $s=({\rm proprio}, {\rm obs}, {\rm goal}, {\rm dist}, {\rm time}) \in \mathbb{R}^{19}$.

\subsubsection{\textbf{Actions}}
Each action is denoted by a vector \mbox{$a \in ([-1, 1])^4$}, which represents (\textit{i}) the initial and (\textit{ii}) the final shoulder joint values, (\textit{iii}) the duration of trajectory execution, and (\textit{iv}) the releasing object time.

\subsubsection{\textbf{Transition function}}
In each training episode, we set the pose of the goal randomly and then, set the pose of the obstacle between the robot and the goal pose with $\pm5cm$ randomness on the x-axis.
Therefore, the transition function is determined by executing the throwing trajectory given the sampled parameters. 
Specifically, the next state,~$s_{i+1}$, can be computed after executing the action~$f_s$; i.e., \mbox{$s_{i+1} = f_s(s_i, a_i)$}, where $s_{i}$ and $a_{i}$ are the current state and action respectively. It should be noted that since the transition function is unknown, our off-policy reinforcement learning framework is also model-free.

\subsubsection{\textbf{Rewards}}
A success is reached if the thrown object falls into the target basket after executing the throwing action. In particular, we calculate the absolute distance between the object and the goal ${\rm dis}({\rm o}, g)<{\rm d}$, where ${\rm d}$ denotes the radius of a cylindrical space that is fitted inside the basket. 

As a result of the current state and the action taken, if the thrown object collides with the obstacle, severe punishment is given by a negative reward $r=-10$. It should be noted that the collision information can only be obtained after the action has been executed. If the next state results in success (the thrown object falls into the basket), we encourage such behavior by setting the reward $r=1$. In the case of throwing the object outside the target basket is also penalized by calculating rewards based on distance $r=-~{\rm dis}({\rm o}, g)$. An episode is terminated after executing a throwing action. It should be noted that successful attempts are recorded for later use in behavioral cloning.
\subsection{Perception}
\begin{figure}[!b]
\centering
\vspace{-5mm}
\includegraphics[width=0.9\linewidth]{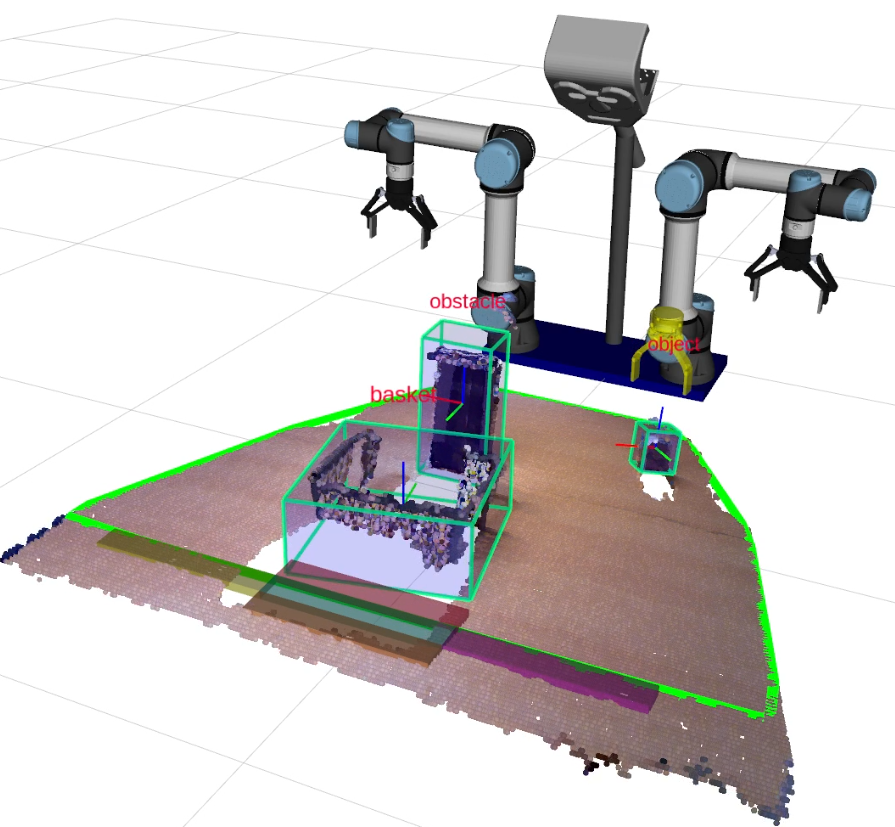}
\caption{Visualizing the output of our perception module for an example scene: it provides world model information in terms of object pose, size, and label. In particular, the estimated object's pose is shown by a reference frame, the object's size is demonstrated by a green bounding box, and the label of the object is highlighted on top of the object's Z axis by red. Moreover, the predicted grasp pose for the target object is shown by a yellow gripper.  }
\label{fig:perception}
\end{figure}
\begin{figure*}[!t]
\centering
\includegraphics[width=0.95\linewidth]{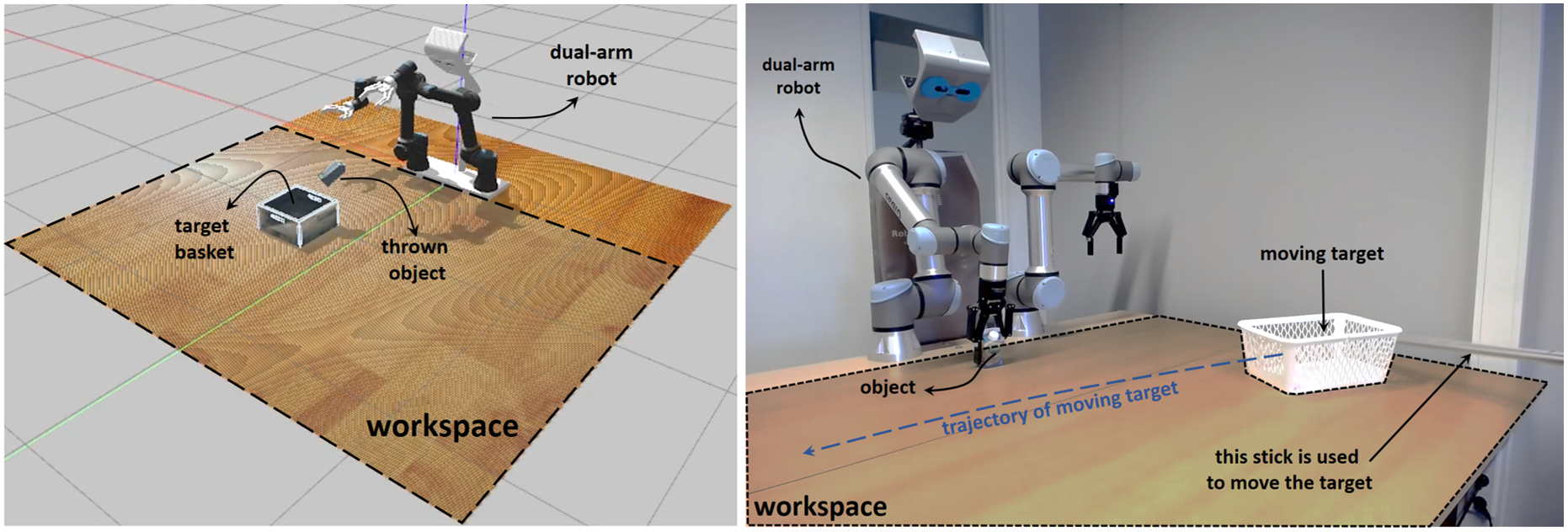}
\caption{Our experimental setups in (\textit{left}) simulation  and (\textit{right}) real-robot settings: our dual-arm robot consists of two UR5e arms for manipulating objects, and an Asus Xiton RGB-D camera to perceive the environment. We have developed a simulation environment in Gazebo very similar to our real robot to reduce the gap between the simulation and the real robot and facilitate transfer learning. Due to safety constraints, we initially trained the robot in the simulation environment, and then directly transferred the learned policy to the real robot without fine-tuning.}
\vspace{-4mm}
\label{fig:simulation_and_real_robot}
\end{figure*}
Due to safety considerations, we trained the proposed throwing policy in simulation first and then use the learned policy on our real-robot platform. In the case of simulation, we developed an interface that provides all necessary information based on Gazebo's services, while for real-robot experiments, we exploit our robotic setup to detect and track the pose of the objects, recognize them, and predict stable grasp syntheses for each of the object in 3D space \cite{kasaei2018towards}\cite{kasaei2019interactive}\cite{kasaei2021simultaneous}. In particular, our real robot uses an RGB-D Asus Xtion camera to perceive the environment by capturing point cloud data at $30$ Hz. Generally, a point cloud consists of a set of points, $p_i : i \in \{1,\dots,n\}$, where each point is described by its 3D coordinates $[x, y, z]$ and RGB-D information. 
To track the pose and speed of the target object, we use a particle filter that considers shape and color data~\cite{kasaei2018perceiving}. It is worth mentioning that for this work, we force the system to grasp the object from above near the center of mass. In particular, our perception system provides a world model service that the agent can call at each time step to receive the current state of the world, which includes the unique ID, pose, speed, label, and grasp synthesis of each object. Figure~\ref{fig:perception} shows the outputs of our perception system regarding object detection (i.e., highlighted by green bounding boxes and reference frames), object recognition (i.e., highlighted by red on top of each object), and pre-grasp pose (i.e., highlighted by yellow gripper). For more information about our perception and grasping pipelines, please refer to our earlier works~\cite{kasaei2018towards}\cite{kasaei2021simultaneous}.





\section{Experiments and Results}
We performed multiple rounds of experiments in both simulation and real-robot settings to validate our method. In this work, we evaluate the performance of the proposed approach based on the \textit{throwing success rate}, which is calculated as the number of times a thrown object lands into the target box divided by the number of trials. More specifically, we tried to investigate the following questions:
(1) Which of the RL approach outperforms other baselines in terms of throwing success rate when used in the same environment and task settings?
(2) Does our method learn to safely throw an object into a target basket while there are obstacles obstructing its path?
(3) Can the policy learned in simulation transfers well to our real-robot where noise and uncertainty exist?

\subsection{Experimental setup and tasks settings}

Our experimental setups in simulation and real-robot are depicted in Fig.~\ref{fig:simulation_and_real_robot}.
In particular, we developed a simulation environment in Gazebo utilizing an ODE physics engine, which is very similar to our real-robot setup. Our setup consists of an Asus xtion camera, two Universal Robots (UR5e) equipped with two-fingered Robotiq 2F-140 gripper, and a user interface to start and stop the experiments. 

Due to safety consideration, we trained the proposed throwing policy in the simulation. After training phase, we conducted experiments in both simulation and real-robot setups to assess the performance of the learned policy in throwing various objects into a box located outside the robot's reachable area. For evaluation purposes, we designed three tasks with ascending difficulty levels: 
\begin{itemize}
    \item \textbf{Task1:} obstacle-free object throwing into a static basket randomly placed in front of the robot. An example of this task in simulation environment is shown in Fig.~\ref{fig:simulation_and_real_robot}~(\textit{left}); 
    \item \textbf{Task2:} obstacle-free object throwing into a moving basket. An example of this task in real-robot setup is shown in Fig.~\ref{fig:simulation_and_real_robot}~(\textit{right});
    \item \textbf{Task3:} object throwing while an obstacle obstructs the throwing path as shown in Fig.~\ref{fig:obstacle_avoidance}. 
\end{itemize}
The dimension of the basket was $0.30 \times 0.30 \times 0.15$ (W $\times$ L $\times$ H) and obstacle was $0.15 \times 0.10 \times 0.22$.
We used $10$ simulated daily-life objects with different materials, shapes, sizes, and weight, and $5$ real objects. In particular, five simulated objects were used during training (i.e., \textit{milk\_box, coke can, banana, bottle, apple}) and the other five simulated objects were used for testing (i.e., \textit{beer\_can, peach, soap, pringles, mustard\_bottle}). For the real robot experiments, we used five household objects that are distinct in size and shape from the simulated objects used during the training phase (i.e., \textit{ugly\_toy, hello\_kitty, small\_box, juice\_box, hand\_soap}).

\begin{figure*}[!t]
\centering
\includegraphics[width=0.95\linewidth, trim={0cm, 0cm, 0cm, 0cm}, clip=true]{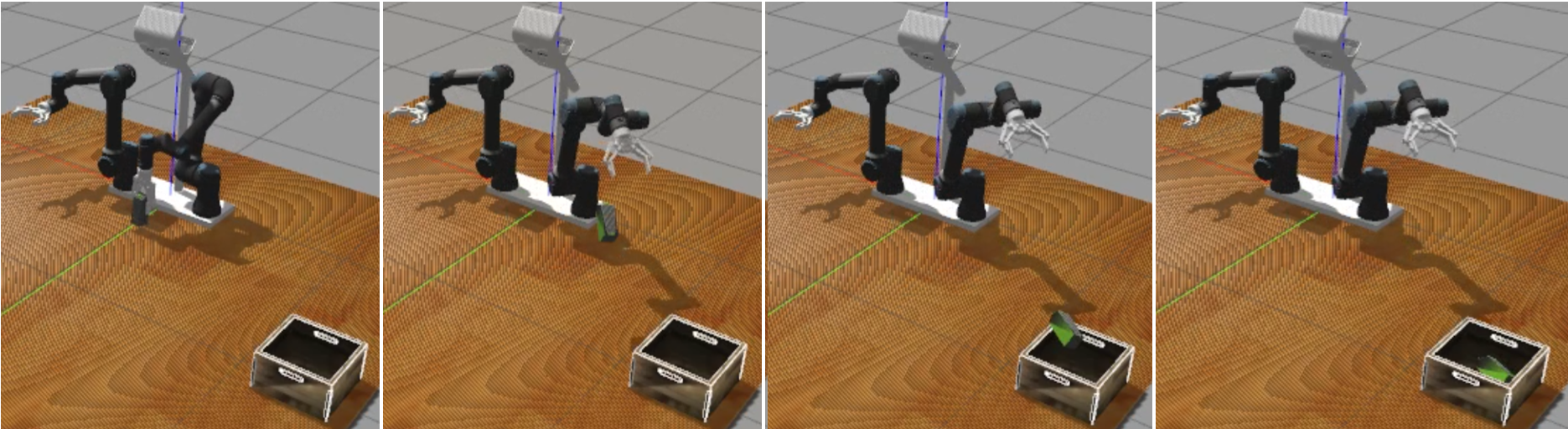}
\caption{Four consecutive snapshots showing our simulated robot successfully thrown a milk box into the basket (\textit{Task1}).In this round of experiments, the basket is not reachable by the robot, and the robot should grasp the target object first, and then, based on the pose of the basket, infers proper parameters to throw the object into the basket successfully.}
\label{fig:seq_tossing_sim}
\vspace{-4mm}
\end{figure*}

\subsection{Baseline Methods}

We employed two sample efficient state-of-the-art off-policy RL algorithms to train the robot: Deep Deterministic Policy Gradient~(DDPG)~\cite{lillicrap2015continuous,silver2014deterministic}, and Soft Actor-Critic~(SAC)~\cite{haarnoja2018soft,haarnoja2018soft2} via stable baseline3~\cite{stable-baselines3}. The architectures of the neural networks for both SAC and DDPG consist of two hidden layers with the size of $256$ neron per layers with the ReLU activation functions. The hyper-parameters of SAC are listed in  Table.~\ref{tab:SAC_hyper}, and the hyper-parameters of DDPG that are not listed in Table ~\ref{tab:SAC_hyper} 
\begin{wraptable}{r}{0.55\linewidth}
    \centering    
    \caption{SAC hyper-parameters}
    \resizebox{0.95\linewidth}{!}{
    \begin{tabular}{|c|c|}
    \hline
    \textbf{Parameter} & \textbf{Value} \\
    \hline
\#hidden layers (all networks) &  $2$ \\
\#hidden units per layer & $256$\\
\#samples per minibatch & $256$\\
optimizer & Adam\\
learning rate  & $3 \times 10^{-4}$\\
batch size & 256 \\
\#epochs & 50K \\
discount ($\gamma$)  & $0.99$\\
replay buffer size &  $10^6$\\
nonlinearity & ReLU\\
target update rate ($\tau$) & 0.005\\
target update interval & 1 \\
gradient steps & 1 \\\hline
    \end{tabular}}
    \label{tab:SAC_hyper}
    \vspace{-3mm}
\end{wraptable}
are reported in the Experiments section of our previous work~\cite{luo2020accelerating}. We also considered Behavior Cloning (BC), which directly learns a policy (i.e., similar to the actor network) by using supervised learning on observation-action pairs from $25$k successful trials, collected during train and test phases.  

\subsection{Results}
For each of the proposed tasks, we trained the model for $50,000$ steps in simulation using the five training objects. In the case of simulation, for each object the robot's throwing performance was tested for $100$ steps using the learned policy twice: once with the five test (unseen) objects and once with the five train (seen) objects. We also used the same learned policy for real robot experiments and test each of the test objects for $20$ times. As opposed to ``unseen objects'', which is a mixed set of objects not seen during training, ``seen objects'' is a mixed set of objects that were used during training. The average throwing success rates for each approach is reported in Table~\ref{tab:results}. 
\begin{figure*}[!t]
    \centering
\includegraphics[width=0.95\linewidth, trim={0cm, 0cm, 0cm, 0cm}, clip=true]{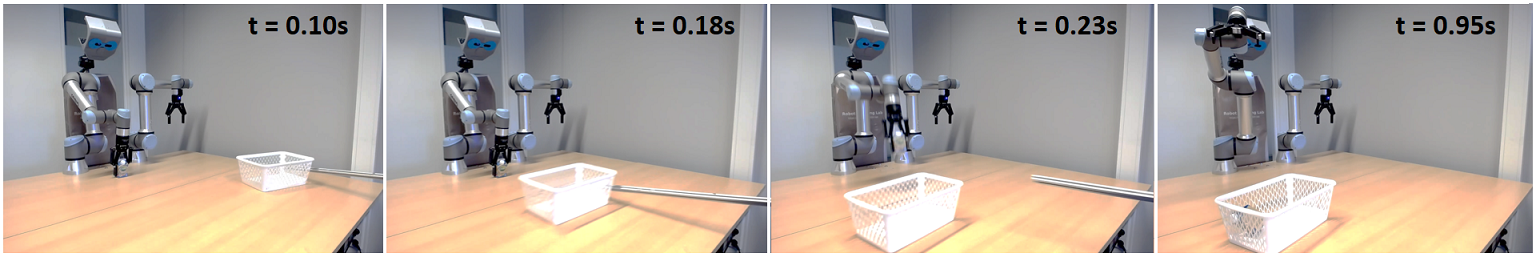}
\caption{A sequence of snapshots showing our simulated robot successfully thrown a milk box into the moving basket (\textit{Task2}): In this round of experiments, the basket is not reachable by the robot, and a human user moves the basket using an aluminum stick. The robot should first estimate the direction and velocity of the basket based on visual information, and then infer proper parameters to throw the milk box into the basket successfully.}
\label{fig:task2}
\vspace{-2mm}
\end{figure*}
\begin{figure*}[!t]
    \centering
    \includegraphics[width=0.95\linewidth]{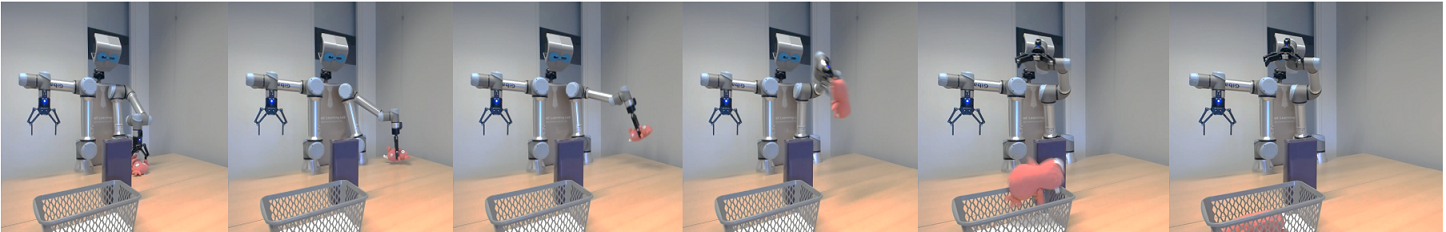}
    \caption{Tossing object into the basket while avoiding the obstacle: the robot should first detect the pose of the object, basket and obstacle, and then infers appropriate parameters to throw the object into the basket without colliding with the obstacle.  }
    \label{fig:task3}
    \vspace{-4mm}
\end{figure*}

In the case of \textbf{Task 1} (i.e., obstacle-free object throwing into a randomly placed static basket), we observed that for the \textit{seen} objects the robot with SAC policy could throw the objects into the basket successfully with $94\%$ accuracy, while with DDPG and BC (on average) achieved $91\%$ and $86\%$ success rate, respectively. A sequence of snapshots demonstrating our robot successfully throwing a milk box into a basket is depicted in Fig.~\ref{fig:seq_tossing_sim}.
Regarding \textit{unseen} objects, the robot, on average, obtained up to $91\%$ ($91$ success out of $100$ attempts) throwing success rate in simulation and $90\%$ ($18/20$) in real-robot setting. In particular, the robot with SAC method, on average, showed the best throwing performance in real and simulation experiments, and outperformed DDPG and BC methods with a large margin. The robot with DDPG achieved the second-best throwing success rate (real: $85\%$ ($17/20$), simulation: $81\%$), whereas using BC approach it could get $70\%$ ($14/20$) and $81\%$ accuracy in real and simulation experiments respectively. As expected, the success rate of throwing unseen objects is moderately lower for all policies. These results showed that the learned policy performs well both in simulation and for the real robot.

\begin{table}[!b]
    \centering
    \vspace{-4mm}
    \caption{Object Throwing Performance (Mean \%). The S/U-S/R shows the setup configuration: the first token refers to the \textbf{S}een or \textbf{U}nseen objects, and the second one denotes \textbf{S}imulation or \textbf{R}eal experiments.}
     \resizebox{\linewidth}{!}{
    \begin{tabular}{|c|c|c|c|c|c|c|c|c|c|}
      \hline
       \multirow{2}{*}{\textbf{Methods}} &\multicolumn{3}{c|}{\textbf{Task 1}}&\multicolumn{3}{c|}{\textbf{Task 2}}&\multicolumn{3}{c|}{\textbf{Task 3}} \\\cline{2-10} 
       & \textbf{S-S} & \textbf{U-S} & \textbf{U-R} & \textbf{S-S} & \textbf{U-S} & \textbf{U-R} & \textbf{S-S} & \textbf{U-S} & \textbf{U-R} \\\hline
       SAC   & \textbf{94} & \textbf{91} & \textbf{90} & \textbf{91} & \textbf{86} & \textbf{85} & \textbf{86} & \textbf{83} & \textbf{80}   \\\hline
       DDPG  & 92 & 81 & 85 & 89 & \textbf{86} & 75 & 85 & 77 & 65  \\\hline
       BC    & 86 & 81 & 70 & 79 & 73 & 65 & 72 & 67 & 55  \\\hline
    \end{tabular}}
    
    \label{tab:results}
\end{table}

In the second round of experiments (\textbf{Task2}) the robot should learn to throw a target object into a moving basket. In the case of simulation experiments, we randomly selected the direction and  a linear speed for moving the basket while in real-robot experiments, a human user moved the basket using a stick (see Fig.~\ref{fig:task2}). Similar to the previous round of experiments, the robot with the SAC policy obtained the best results for both seen and unseen objects. More specifically, for the \textit{seen} objects, the robot obtained $91\%$ with the SAC strategy, whereas its throwing performance with DDPG and BC dropped to $89\%$ and $79\%$, respectively. When tossing unseen objects in simulation, the robot was $86\%$ accurate with SAC and DDPG policies, compared to $73\%$ accuracy with BC. Intriguingly, in contrast to simulation results, the robot with DDPG policy does not perform as competitively to SAC in the real-world.

\textbf{Task 3} is much more complex than the previous tasks as the robot should learn to infer an obstacle-free path to throw the object into the basket. An example of such expriments is shown in Fig.~\ref{fig:task3}.
Similar to the previous round of experiments, the robot with SAC policy outperformed DDPG and BC for both \textit{seen} and \textit{unseen} objects. By comparing all experiments, we observed that the difference between SAC policy and others is larger when the task is more difficult. This becomes more visible in the case of unseen objects in real-robot experiments. In particular, the robot with SAC policy could better model the solution space and handle unmodeled physical parameters presented in real-world objects (e.g., aerodynamics, materials, or stiffness). Therefore, in simulation experiments, the robot using SAC policy obtained $86\%$ and $83\%$ throwing accuracy for seen and unseen objects respectively, while in real-robot experiments it could achieve $80\%$ success rate. While the performance of the robot with DDPG policy was marginally lower than SAC policy for seen objects (i.e., 86\% vs. 85\%), the difference increased significantly for unseen objects (i.e., simulation 83\% vs. 77\%, and in real 80\% vs 65\%). Experimental results showed that our formulation maintains the flexibility needed to express complicated dynamics system while also making learning easier through trial and error.

\subsection{Failure Cases}

The main reason for failure for all approaches in the simulation experiments was the inaccurate parameters prediction and throwing of the object near the basket. In the case of real robot experiments, apart from inaccurate parameters prediction, we observed three other type of failures: (\textit{i}) \textit{Inaccurate Tracking} was one of the main reasons, where the user move the object really fast and the tracking could not follow the pose of the object immediately; (\textit{ii}) the second primary reason was the lag in executing the gripper commands on time. In particular,  the gripper is controlled through the robot's controller and we can not control it directly, i.e., we need to send the command to the robot's controller first, and then the controller sends the command to the gripper, which depends on the status of network and the robot. (\textit{iii}) Selecting an unstable grasp pose was the third reason for failure. Furthermore, we found that for objects with heterogeneous shapes like \textit{ugly\_toy}, \textit{hand\_soap}, and \textit{hello\_kitty}, the trajectory of the thrown object varied depending on the grasp pose. In contrast, for homogeneous objects, the trajectory of the thrown object was not dependent on the grasp pose.

\vspace{-1mm}
\section{CONCLUSIONS}

In this paper, we trained a policy to modulate throwing kernel parameters through RL to enable robots to precisely throw objects into a moving basket while there are obstacles obstructing the path. In particular, our method learns to handle a wide range of situations and avoid colliding the thrown object with the obstacle.
With our formulation, the robot could iteratively learn the aspects of dynamics that are difficult to model analytically. Due to safely constraints, we trained the throwing policy in simulation and directly applied the learned policy in real-robot. We performed extensive sets of experiments in both simulation and real robot setups in three scenarios with ascending level of difficulties including, tossing objects into a (\textit{i}) static basket, (\textit{ii}) moving basket, and (\textit{iii}) obstacle avoidance. Experimental results showed that the proposed approach enables robot to precisely throw the object into the basket. We also observed that the learnt policy could be directly applied to the real robot even though it had only been trained in simulation. It also showed outstanding generalization capabilities to new target locations and unknown objects. In the continuation of this work, we would like to investigate the possibility of enhancing throwing performance by taking into account additional sensory modalities (e.g., force or tactile sensors), as these could help the robot grasp objects steadily and adjust its throwing parameters accordingly.

{
\small
\bibliographystyle{IEEEtran}
\bibliography{refs}

\begin{thebibliography}{10}
\providecommand{\url}[1]{#1}
\csname url@samestyle\endcsname
\providecommand{\newblock}{\relax}
\providecommand{\bibinfo}[2]{#2}
\providecommand{\BIBentrySTDinterwordspacing}{\spaceskip=0pt\relax}
\providecommand{\BIBentryALTinterwordstretchfactor}{4}
\providecommand{\BIBentryALTinterwordspacing}{\spaceskip=\fontdimen2\font plus
\BIBentryALTinterwordstretchfactor\fontdimen3\font minus
  \fontdimen4\font\relax}
\providecommand{\BIBforeignlanguage}[2]{{%
\expandafter\ifx\csname l@#1\endcsname\relax
\typeout{** WARNING: IEEEtran.bst: No hyphenation pattern has been}%
\typeout{** loaded for the language `#1'. Using the pattern for}%
\typeout{** the default language instead.}%
\else
\language=\csname l@#1\endcsname
\fi
#2}}
\providecommand{\BIBdecl}{\relax}
\BIBdecl

\bibitem{kuhn2016throwing}
J.~E. Kuhn, ``Throwing, the shoulder, and human evolution.'' \emph{American
  Journal of Orthopedics (Belle Mead, NJ)}, vol.~45, no.~3, pp. 110--114, 2016.

\bibitem{hu2010ball}
J.-S. Hu, M.-C. Chien, Y.-J. Chang, S.-H. Su, and C.-Y. Kai, ``A ball-throwing
  robot with visual feedback,'' in \emph{2010 IEEE/RSJ International Conference
  on Intelligent Robots and Systems}.\hskip 1em plus 0.5em minus 0.4em\relax
  IEEE, 2010, pp. 2511--2512.

\bibitem{gai2013motion}
Y.~Gai, Y.~Kobayashi, Y.~Hoshino, and T.~Emaru, ``Motion control of a ball
  throwing robot with a flexible robotic arm,'' \emph{International Journal of
  Computer and Information Engineering}, vol.~7, no.~7, pp. 937--945, 2013.

\bibitem{zeng2020tossingbot}
A.~Zeng, S.~Song, J.~Lee, A.~Rodriguez, and T.~Funkhouser, ``Tossingbot:
  Learning to throw arbitrary objects with residual physics,'' \emph{IEEE
  Transactions on Robotics}, vol.~36, no.~4, pp. 1307--1319, 2020.

\bibitem{takahashi2021dynamic}
A.~Takahashi, M.~Sato, and A.~Namiki, ``Dynamic compensation in throwing motion
  with high-speed robot hand-arm,'' in \emph{2021 IEEE International Conference
  on Robotics and Automation (ICRA)}.\hskip 1em plus 0.5em minus 0.4em\relax
  IEEE, 2021, pp. 6287--6292.

\bibitem{ploeger2022controlling}
K.~Ploeger and J.~Peters, ``Controlling the cascade: Kinematic planning for
  n-ball toss juggling,'' \emph{arXiv preprint arXiv:2207.01414}, 2022.

\bibitem{mason1993dynamic}
M.~T. Mason and K.~M. Lynch, ``Dynamic manipulation,'' in \emph{Proceedings of
  1993 IEEE/RSJ International Conference on Intelligent Robots and Systems
  (IROS'93)}, vol.~1.\hskip 1em plus 0.5em minus 0.4em\relax IEEE, 1993, pp.
  152--159.

\bibitem{kober2012learning}
J.~Kober, K.~Muelling, and J.~Peters, ``Learning throwing and catching
  skills,'' in \emph{2012 IEEE/RSJ International Conference on Intelligent
  Robots and Systems}.\hskip 1em plus 0.5em minus 0.4em\relax IEEE, 2012, pp.
  5167--5168.

\bibitem{lofaro2012humanoid}
D.~M. Lofaro, R.~Ellenberg, P.~Oh, and J.-H. Oh, ``Humanoid throwing: Design of
  collision-free trajectories with sparse reachable maps,'' in \emph{2012
  IEEE/RSJ International Conference on Intelligent Robots and Systems}.\hskip
  1em plus 0.5em minus 0.4em\relax IEEE, 2012, pp. 1519--1524.

\bibitem{ghadirzadehdeep}
A.~Ghadirzadeh, A.~Maki, D.~Kragic, and M.~Bj{\"o}rkman, ``Deep predictive
  policy training using reinforcement learning. in 2017 ieee,'' in \emph{RSJ
  International Conference on Intelligent Robots and Systems (IROS)}, pp.
  2351--2358.

\bibitem{kober2011reinforcement}
J.~Kober, E.~Oztop, and J.~Peters, ``Reinforcement learning to adjust robot
  movements to new situations,'' in \emph{Twenty-Second International Joint
  Conference on Artificial Intelligence}, 2011.

\bibitem{kasaei2018towards}
S.~H. Kasaei, M.~Oliveira, G.~H. Lim, L.~S. Lopes, and A.~M. Tom{\'e},
  ``Towards lifelong assistive robotics: A tight coupling between object
  perception and manipulation,'' \emph{Neurocomputing}, vol. 291, pp. 151--166,
  2018.

\bibitem{kasaei2019interactive}
S.~H. Kasaei, N.~Shafii, L.~S. Lopes, and A.~M. Tom{\'e}, ``Interactive
  open-ended object, affordance and grasp learning for robotic manipulation,''
  in \emph{2019 International Conference on Robotics and Automation
  (ICRA)}.\hskip 1em plus 0.5em minus 0.4em\relax IEEE, 2019, pp. 3747--3753.

\bibitem{kasaei2021simultaneous}
H.~Kasaei, S.~Luo, R.~Sasso, and M.~Kasaei, ``Simultaneous multi-view object
  recognition and grasping in open-ended domains,'' \emph{arXiv preprint
  arXiv:2106.01866}, 2021.

\bibitem{kasaei2018perceiving}
S.~Kasaei, J.~Sock, L.~S. Lopes, A.~M. Tom{\'e}, and T.-K. Kim, ``Perceiving,
  learning, and recognizing 3d objects: An approach to cognitive service
  robots,'' in \emph{Proceedings of the AAAI Conference on Artificial
  Intelligence}, vol.~32, no.~1, 2018.

\bibitem{lillicrap2015continuous}
T.~P. Lillicrap, J.~J. Hunt, A.~Pritzel, N.~Heess, T.~Erez, Y.~Tassa,
  D.~Silver, and D.~Wierstra, ``Continuous control with deep reinforcement
  learning,'' \emph{arXiv preprint arXiv:1509.02971}, 2015.

\bibitem{silver2014deterministic}
D.~Silver, G.~Lever, N.~Heess, T.~Degris, D.~Wierstra, and M.~Riedmiller,
  ``Deterministic policy gradient algorithms,'' in \emph{International
  conference on machine learning}.\hskip 1em plus 0.5em minus 0.4em\relax PMLR,
  2014, pp. 387--395.

\bibitem{haarnoja2018soft}
T.~Haarnoja, A.~Zhou, P.~Abbeel, and S.~Levine, ``Soft actor-critic: Off-policy
  maximum entropy deep reinforcement learning with a stochastic actor,'' in
  \emph{International conference on machine learning}.\hskip 1em plus 0.5em
  minus 0.4em\relax PMLR, 2018, pp. 1861--1870.

\bibitem{haarnoja2018soft2}
T.~Haarnoja, A.~Zhou, K.~Hartikainen, G.~Tucker, S.~Ha, J.~Tan, V.~Kumar,
  H.~Zhu, A.~Gupta, P.~Abbeel \emph{et~al.}, ``Soft actor-critic algorithms and
  applications,'' \emph{arXiv preprint arXiv:1812.05905}, 2018.

\bibitem{stable-baselines3}
\BIBentryALTinterwordspacing
A.~Raffin, A.~Hill, A.~Gleave, A.~Kanervisto, M.~Ernestus, and N.~Dormann,
  ``Stable-baselines3: Reliable reinforcement learning implementations,''
  \emph{Journal of Machine Learning Research}, vol.~22, no. 268, pp. 1--8,
  2021. [Online]. Available: \url{http://jmlr.org/papers/v22/20-1364.html}
\BIBentrySTDinterwordspacing

\bibitem{luo2020accelerating}
S.~Luo, H.~Kasaei, and L.~Schomaker, ``Accelerating reinforcement learning for
  reaching using continuous curriculum learning,'' in \emph{2020 International
  Joint Conference on Neural Networks (IJCNN)}.\hskip 1em plus 0.5em minus
  0.4em\relax IEEE, 2020, pp. 1--8.

\end{thebibliography}
}

\end{document}